%% file: main.tex
\documentclass[11pt]{article}

\usepackage[table]{xcolor}
\usepackage[preprint]{acl}
\usepackage{times}
\usepackage{latexsym}
\usepackage[T1]{fontenc}
\usepackage[utf8]{inputenc}
\usepackage{microtype}
\usepackage{inconsolata}
\usepackage{tabularx}
\usepackage{array}
\usepackage{subcaption}
\newcolumntype{Y}{>{\raggedleft\arraybackslash}X}

\usepackage{graphicx}
\usepackage{booktabs}
\usepackage{amsmath}
\definecolor{switchblue}{RGB}{232, 243, 255}
\input{math_commands.tex}

\usepackage[most]{tcolorbox}

\definecolor{memoryblue}{HTML}{2563A6}

\newtcolorbox{switchmemprompt}[1]{
  enhanced,
  colback=white,
  colframe=black!45,
  colbacktitle=black!6,
  coltitle=black,
  fonttitle=\bfseries,
  title=#1,
  boxrule=0.5pt,
  arc=1mm,
  left=1.4mm,
  right=1.4mm,
  top=1.2mm,
  bottom=1.2mm,
  before skip=0pt,
  after skip=0pt,
}
\newcommand{\memorylabel}[1]{%
  \textcolor{memoryblue}{\bfseries #1}\par}

\title{Learning How to Remember: Memory Modality Selection for GUI Agents}

\hypersetup{
  pdftitle={Learning How to Remember: Memory Modality Selection for GUI Agents},
}
\author{
    \textbf{Taeyun Roh}$^{1}$ \quad
    \textbf{Suhyeong Park}$^{1}$ \quad
    \textbf{Gijin Kim}$^{1}$ \\
    \textbf{Wonjune Jang}$^{2}$ \quad
    \textbf{Jaewoo Kang}$^{1,3}$ \\
    $^{1}$Korea University \quad
    $^{2}$Myongji University \quad
    $^{3}$AIGEN Sciences \\
    \texttt{\{nrbsld,kangj\}@korea.ac.kr}
}

\begin{document}

\maketitle

\begin{abstract}
Recent advances in vision-language models (VLMs) have substantially improved the capabilities of GUI agents, while memory augmentation has emerged as a promising approach for reusing knowledge from past interactions. 
However, existing memory augmented GUI agents typically adopt a fixed representation strategy, providing retrieved experience as a static modality. It remains unclear whether a single memory modality is consistently optimal, or how modality preference changes across interaction steps. In this work, we conduct a systematic study of memory modality in VLM based GUI agents. 
Across GUI benchmarks, we find that the preferred memory representation is strongly step dependent. Motivated by the finding, we introduce \textbf{SwitchMem}, a routing framework that dynamically selects between textual and visual memory representations at each step.
Across three GUI benchmarks, SwitchMem outperforms memory augmented baselines that rely on static memory representations. Our findings validate our claim that memory modality is a step dependent decision rather than a fixed design choice. Code and project resources are available
\href{https://txxnrd.github.io/switchmem}
{\texttt{here}}.
\end{abstract}

\section{Introduction}

Recent advances in vision-language models (VLMs) have substantially improved
their ability to perceive, reason over, and act upon visual environments
~\citep{xie2024osworld,team2025gemini,jung2026breaking}.
Such progress has enabled graphical user interface (GUI) agents, where VLMs directly interpret screenshots, reason about user instructions, and execute actions across desktop and mobile environments
~\citep{xu2024aguvis,zhang2025appagent,rawles2025androidworld,qin2025ui}.
Despite these advances, GUI interaction remains challenging: tasks often
require long-horizon decision making, and errors made at one step can
propagate to subsequent decisions.

To learn from prior interactions and reduce repeated errors, recent GUI agents augment the actor with memories constructed from past trajectories, demonstrations, interaction knowledge, and visual observations ~\citep{li2026echotrail,qin2026executable,choi2026naive,
liu-etal-2026-learnact,zeng2026mementogui}.
However, existing systems generally adopt predefined memory representations, presenting retrieved experience solely as text, visual observations, or a fixed combination of both modalities.
Once chosen, this representation strategy typically remains unchanged across interaction steps.

\begin{figure}[t]
    \centering
    \includegraphics[width=\linewidth]{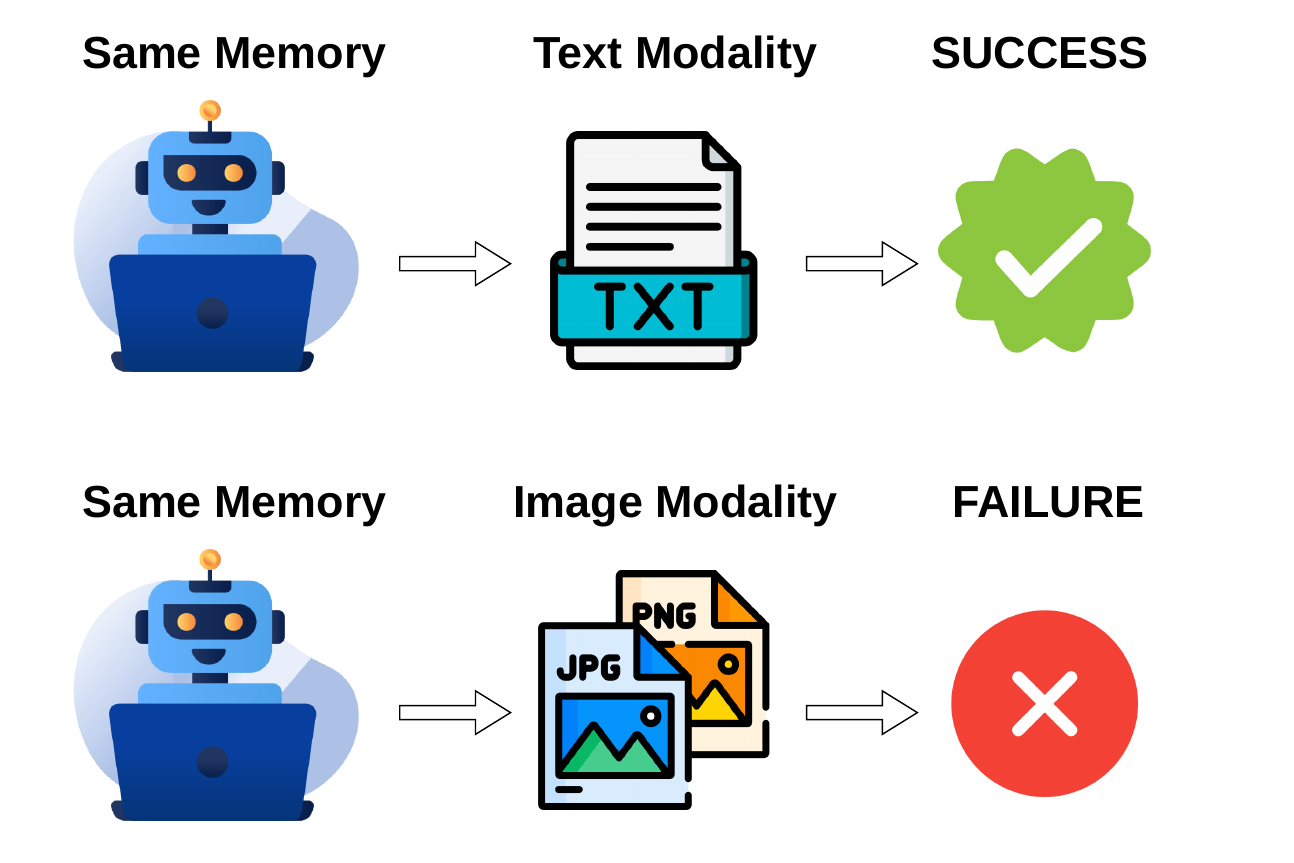}
    \caption{
    The same retrieved memory can lead to different outcome for a GUI agent depending on
    modality.
    }
    \label{fig:motivation}
\end{figure}
However, existing GUI memory systems rarely question whether a single representation modality is suitable for every interaction step. Prior work on VLMs suggests that textual and visual representations provide
fundamentally different forms of information and that their relative utility
depends on the reasoning context.~\citep{pezeshkpour2025mixed,yang2026captionqa,long2026understanding,roh2026buzzycontrastivescoringmitigate}.
\begin{figure}[t]
    \centering
    \includegraphics[width=\linewidth]
{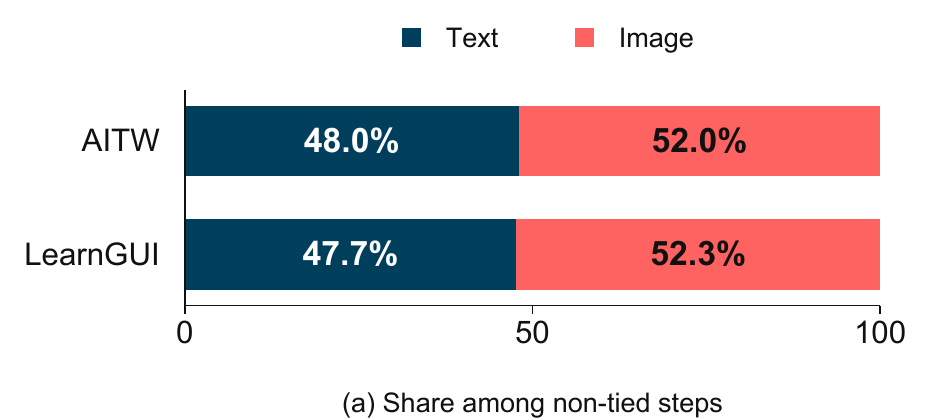}
    \caption{
    Relative frequency of Text and Image wins among interaction steps. Neither modality consistently dominates across benchmarks.
    }
    \label{fig:step-preference}
\end{figure}
Text can provide a compact semantic abstraction of visual content, but
converting images into language may discard fine-grained information required
for downstream reasoning~\citep{yang2026captionqa}.
Conversely, visual representations preserve fine-grained spatial and
appearance information that textual surrogates may fail to retain
~\citep{yang2026captionqa,jenkins2026thinking}. However, when the information needed for reasoning can be captured in structured text, textual abstraction can improve performance on complex visual reasoning tasks~\citep{liu-etal-2023-deplot,park2025generalizing,xu2026vision}. 
Although these findings highlight the context-dependent role of modality in VLM reasoning, it remains unclear whether the preferred memory modality varies across interaction steps in memory enhanced GUI agents.

In this work, we conduct a systematic study of memory modality in VLM based GUI agents.
We show that preferred memory modality changes each step, and that the preference is revealed through models internal hidden states. Based on the observations, we introduce \textbf{SwitchMem}, a lightweight  router that selects between textual and visual memories at each step.
Our contributions are threefold:
\begin{itemize}
    \item We show that the preferred memory modality varies across interaction steps and that fixed modalities lead to suboptimal performance across evaluation settings.
   
    \item We show that memory modality preference is reflected at the
    representation level within the agent.
    Our \emph{Directional Chain-of-Embedding (D-CoE)} analysis reveals a
    reproducible Text/Image preference signal that generalizes to unseen GUI trajectories.
    
    \item Building on this representation-level signal, we propose
    \textbf{SwitchMem}, a routing framework that adaptively selects between Text and Image memory at each interaction step. Consequently, it consistently outperforms all static memory baselines across evaluation tasks.
\end{itemize}


\section{Related Work}

\paragraph{Mobile GUI Agents.}
Mobile GUI agents execute user instructions through repeated observation and interaction with mobile applications.
At each step, the agent interprets the current screenshot in the context of the task instruction and interaction history, then predicts an action such as tapping, swiping, or entering text~\citep{wang2024mobile,wang2025mobile,qin2025ui}.
The resulting interface state becomes the observation for the next decision, and this process continues until the task is completed.
Recent VLM based approaches integrate visual perception, action grounding, and sequential reasoning to support this interaction loop~\citep{xu2024aguvis,qin2025ui}. Despite these advances, reliable mobile interaction remains challenging because each action changes the interface state, and errors in perception or action selection can propagate through subsequent decisions.

\paragraph{Memory Systems for GUI Agents.}
Memory augmentation enables agents to extract and reuse actionable knowledge
from prior interactions, supporting informed decisions and reducing the risk of repeated failures 
\citep{chhikara2025mem0,yeo2026worldmm,xu2026mem,roh-etal-2026-clag}. In GUI environments, existing approaches preserve episodic histories and retrieve relevant experiences as in-context examples~\citep{agashe2025agent,xu2025retrieval,zhang2025appagent,liu-etal-2026-learnact}. Others transform interaction trajectories into reusable textual traces, structured procedures, or executable knowledge that guides action generation ~\citep{li2026echotrail,qin2026executable,xiao2026ui}. Visual memory methods instead retain screenshots or action-relevant regions to ground decisions in prior interface states~\citep{rawles2025androidworld,choi2026naive,zeng2026mementogui}.
While these methods differ in how memories are constructed and retrieved, they use a fixed modality after retrieval. SwitchMem instead treats memory modality as a step level decision, selecting between Text and Image based on the current interaction state.

\section{Understanding Memory Modality Preference}
\label{sec:understanding}

\subsection{Study Setup}
\label{sec:setup}
We compare textual and visual representations of the same retrieved demonstration at each interaction step. By holding the retrieved demonstration fixed, we examine how modality preference varies across steps. We use Qwen3.5-9B as the actor in all primary comparisons.
At step $i$, let
\begin{equation}
    x_i=(g_i,o_i)
\end{equation}
denote the current GUI state, where $g_i$ is the task instruction and $o_i$
is the current screenshot.
Each memory candidate is retrieved from a bank of valid demonstration steps
containing a past GUI state and its successful demonstrated action. To isolate the effect of memory representation from retrieval quality, we use
a shared text-based retriever for both modalities.
For each state $x_i$, the retriever selects a single demonstration $j_i$, which
is then instantiated in either textual or visual form:
\begin{equation}
    m_i^{\mathrm{T}}
    = \bigl(g_{j_i}, t_{j_i}, a_{j_i}\bigr),
    \qquad
    m_i^{\mathrm{I}}
    = \bigl(g_{j_i}, v_{j_i}, a_{j_i}\bigr).
    \label{eq:memory-modalities}
\end{equation}
where $g_{j_i}$ is the task instruction associated with the retrieved
demonstration, $t_{j_i}$ is the textual description of its GUI state,
$v_{j_i}$ is the corresponding screenshot, and $a_{j_i}$ is the
demonstrated action. The textual state representation is constructed from structured UI information provided by each benchmark.

\begin{figure}[t]
    \centering
    \includegraphics[width=\linewidth]{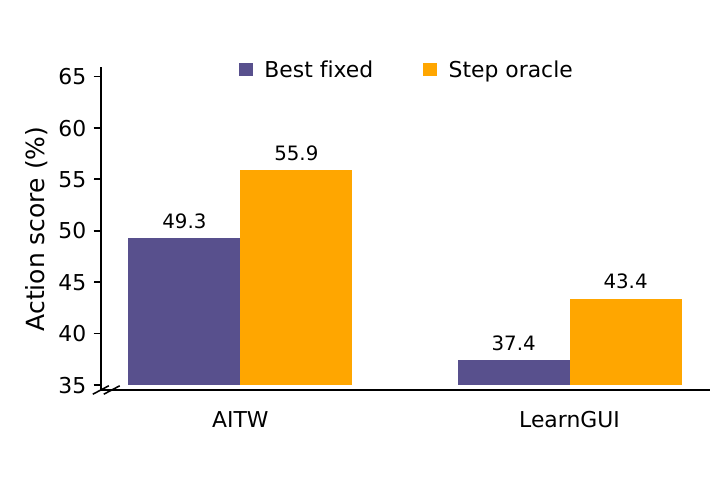}
    
    \caption{Performance gap between fixed modality and step-wise oracle modality memory selection.}
    \label{fig:fixed-oracle-gap}
\end{figure}
At each interaction step $i$, we compare the expected action utility obtained
under Text and Image memory.
For modality $z\in\{\mathrm{T},\mathrm{I}\}$, we estimate its utility using
$K$ sampled actions from the frozen actor:

\begin{equation}
    \hat{u}_i^z
    =
    \frac{1}{K}
    \sum_{k=1}^{K}
    S_b\!\left(a_{i,k}^z, a_i^*\right),
\end{equation}

where $S_b$ is the benchmark specific action scorer and $a_i^*$ is the
ground-truth action.
We define the step-wise memory modality preference as

\begin{equation}
    \hat{\Delta}_i
    =
    \hat{u}_i^{\mathrm{I}}
    -
    \hat{u}_i^{\mathrm{T}}.
    \label{eq:empirical-preference-gap}
\end{equation}

\subsection{Step-wise Memory Modality Preference}
\label{sec:step-wise-preference}
We examine how memory modality preference varies across interaction steps using the estimated utility gap $\hat{\Delta}_i$ defined in Equation~\ref{eq:empirical-preference-gap}.

\paragraph{Neither modality consistently dominates.}
As shown in Figure~\ref{fig:step-preference}, both Text and Image memory are preferred for
substantial subsets of interaction steps across AITW and LearnGUI.
 Image memory yields a slightly higher average performance on both AITW and LearnGUI, but the weaker modality on aggregate still provides higher utility on numerous individual execution steps. The preferred modality therefore depends on the current interaction state.


\begin{figure}[t]
    \centering
    \includegraphics[width=0.95\linewidth]{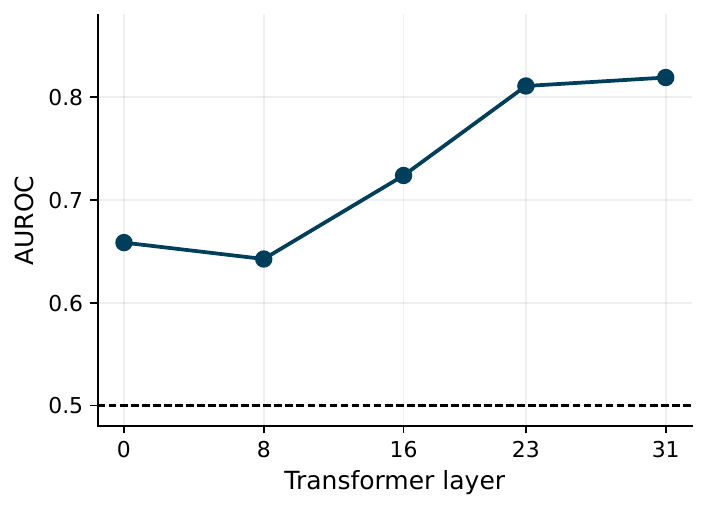}
    \caption{
    layer-wise modality-preference probing with D-CoE on AITW.
    Using examples with at least five-of-seven agreement in sampled preference,
    AUROC increases in deeper layers, indicating a stronger modality-preference
    signal in later hidden representations.
    }
    \label{fig:dcoe-probe}
\end{figure}

\paragraph{Fixed modalities leave consistent performance on the table.}
The step-wise variation in modality preference implies that a single fixed
representation cannot be optimal throughout an entire interaction.
To quantify this effect, we define a step-wise oracle. At each evaluation step, we run the frozen actor separately with Text and Image memory under the same fixed evaluation seed. We score both predicted actions using the benchmark specific metric, and the oracle takes the higher of the two scores.
\begin{figure*}[t]
    \centering
    \includegraphics[width=\linewidth]
    {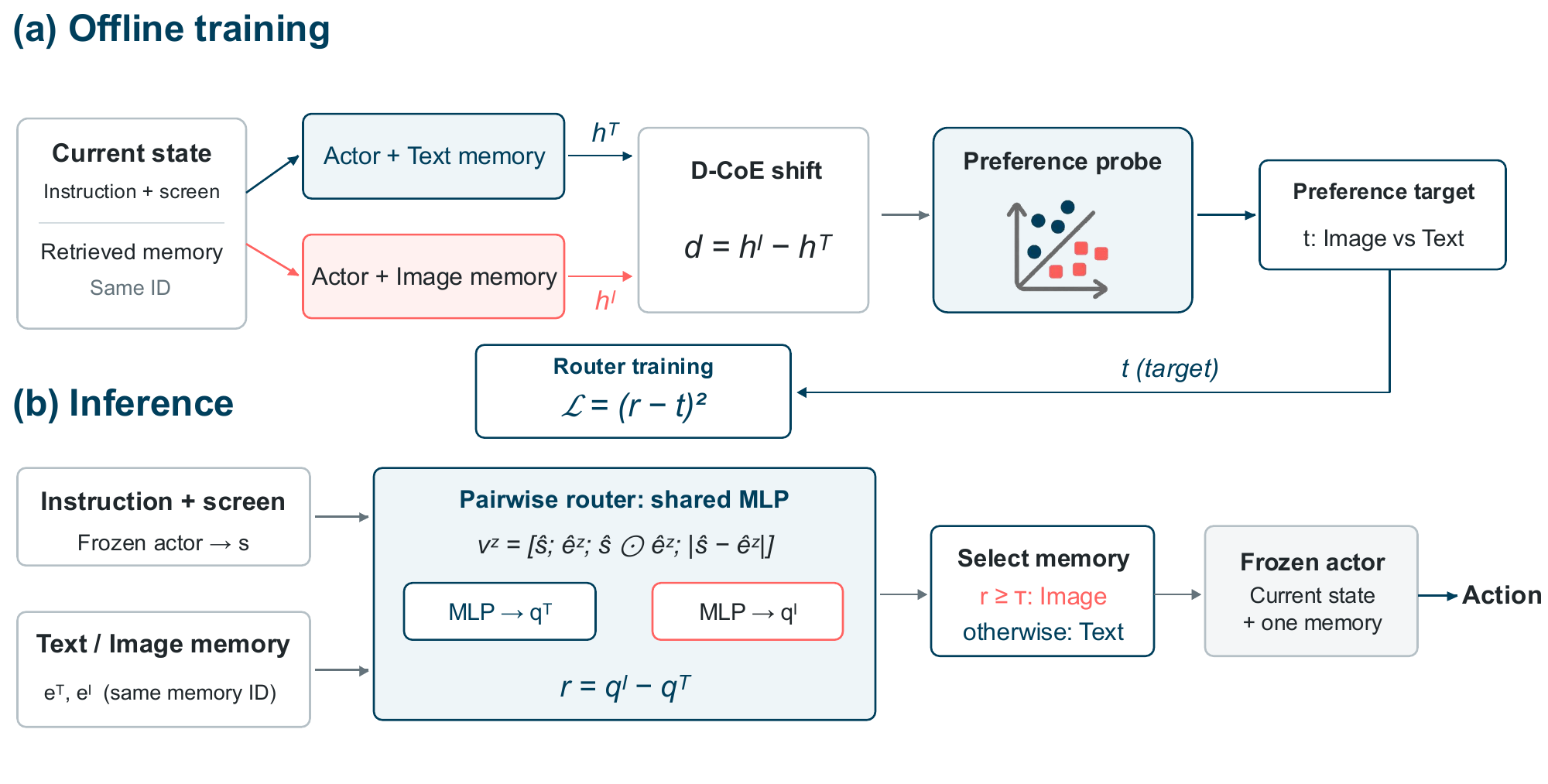}
\caption{
Overview of SwitchMem.
During offline training, the difference between the actor's hidden states
obtained with Image and Text memory is used to construct a continuous
modality preference target. A pairwise router scores both memory candidates
and is trained to match their score difference to this target.
At inference, the router selects one memory modality for the frozen actor
to generate the next action.
}
    \label{fig:main_figure}
\end{figure*}
As shown in Figure~\ref{fig:fixed-oracle-gap}, the step-wise oracle consistently
outperforms the strongest fixed-modality strategy across benchmarks,
revealing a persistent gap caused by static memory representation.
This gap provides direct motivation for adaptive, step-level modality selection.

\subsection{Modality Preference in Hidden States}
\label{sec:modal_preference}
Building on the finding that memory modality preference varies across interaction steps, we probe the actor's hidden representations to determine whether the optimal modality can be predicted at each step.

\paragraph{Directional Chain-of-Embedding.}
Chain-of-Embedding (CoE) analyzes how different input conditions diverge
throughout a VLM by comparing their intermediate hidden representations
~\citep{long2026understanding}.
We adapt this perspective to contrast the retrieved Text and Image memory
conditions.
Let $h_{i,\ell}^{z}$ denote the hidden state of the frozen actor at layer
$\ell$ under memory condition $z\in\{\mathrm{T},\mathrm{I}\}$.
The actor input contains the current task instruction, the current GUI
screenshot, and the retrieved memory represented in modality $z$. Retrieval is performed once based on a text embedding, and the same retrieved
demonstration ID is used under both Text and Image memories.
We extract $h_{i,\ell}^{z}$ from the final input token of the fully constructed
prompt, immediately before action generation.
We retain the signed shift
\begin{equation}
d_{i,\ell}
=
h_{i,\ell}^{\mathrm{I}}
-
h_{i,\ell}^{\mathrm{T}},
\label{eq:dcoe}
\end{equation}
and refer to the layer-wise sequence
$\mathcal{C}_i = \{d_{i,\ell}\}_{\ell=1}^{L}$ as the
\emph{Directional Chain-of-Embedding (D-CoE)}.

\paragraph{Modality preference reflected in hidden representations.}
We probe whether $d_{i,\ell}$ contains information about which memory modality yields higher action utility.
For each transformer layer, we reduce the shift representation with PCA and train a linear probe to predict the preferred modality.
To reduce noise from stochastic action sampling, we retain only examples for which the same modality is preferred in at least five of seven sampled generations.
As shown in Figure~\ref{fig:dcoe-probe}, the modality preference is predictable from the representation shift on AITW.
This suggests that modality preference becomes reflected in the actor's hidden representations.







\section{SwitchMem}
\label{sec:switchmem}
SwitchMem is trained in two stages. First, a preference estimator processes Text and Image memory through separate actor forward passes and maps the resulting representation shift to a preference target. Second, a pairwise router is trained to predict the preference target from the current state and the retrieved memory candidates. The router applies a shared MLP scorer to each state–memory pair and uses the difference between the Text and Image scores as its routing score.
\begin{table*}[t]
\centering
\small
\setlength{\tabcolsep}{7pt}
\renewcommand{\arraystretch}{1.05}

\begin{tabularx}{\textwidth}{@{} l X Y Y Y @{}}
\toprule
Method
& Augmentation
& AITW (\%)
& LearnGUI (\%)
& MobileMiniWoB (\%) \\
\midrule

\multicolumn{5}{@{}l@{}}{\textit{General-purpose}} \\

GPT-5.4-mini
& --
& 34.000 & 24.912 & 51.087 \\

Qwen3.5-9B
& --
& 41.920 & 23.485 & 58.696 \\

\midrule
\multicolumn{5}{@{}l@{}}{\textit{GUI specialized}} \\

Aguvis-7B
& --
& 33.240 & 24.736 & 29.348 \\

UI-TARS-2B-SFT
& --
& 23.260 & 9.365 & 8.696 \\

UI-TARS-7B-SFT
& --
& 29.980 & 25.705 & 23.913 \\

UI-TARS-1.5-7B
& --
& 36.000 & 6.019 & 38.043 \\

\midrule
\multicolumn{5}{@{}l@{}}{\textit{Qwen3.5-9B Memory augmented}} \\

RAG-GUI
& Generated guidelines
& 18.300 & 24.186 & 9.783 \\

M3A
& Accessibility / SoM
& 17.200 & 20.899 & 41.304 \\

AppAgent
& Operation document
& 14.780 & 31.104 & 25.000 \\

Text
& Text memory
& 48.760 & 35.638 & 53.261 \\

Image
& Image memory
& 49.300 & 37.441 & 57.609 \\

Both
& Text and Image 
& 47.080 & 26.919 & 60.870 \\

EchoTrail
& Retrieved text step
& 45.900 & 27.286 & 55.435 \\

AGMem
& Action region crop
& 45.380 & 28.366 & 52.174 \\

\rowcolor{switchblue}
\textbf{SwitchMem}
& \textbf{Adaptive memory}
& \textbf{51.420}
& \textbf{38.131}
& \textbf{63.043} \\

\bottomrule

\end{tabularx}

\caption{
Performance comparison on AITW, LearnGUI, and MobileMiniWoB++.
AITW and LearnGUI use the official benchmark-provided step level action
scores, whereas MobileMiniWoB++ reports task success rate. The memory-augmented group compares different augmentation
strategies using the same frozen Qwen3.5-9B actor and evaluation protocol.
}
\label{tab:results}
\end{table*}
\subsection{Preference Target Construction}
\label{sec:preference-target}
In this phase, the actor's modality dependent representation shift is transformed into a continuous preference target used to train the pairwise router. We use the
final-layer D-CoE representation defined in Equation~\ref{eq:dcoe}:
We select the final transformer block based on the layer wise probing analysis
in Section~\ref{sec:modal_preference}.
To construct the preference targets, we compare the sampled action utilities
obtained with Text and Image memory. We retain the non-tied training states

$$
\mathcal{D}_{\mathrm{pref}}
= \{i \mid \hat{u}_i^{I} \neq \hat{u}_i^{T}\}
$$

and fit a preference estimator to predict whether
$\hat{u}_i^{I} - \hat{u}_i^{T} > 0$.
To reduce sample-specific noise in the 4096-dimensional representation and obtain stable preference targets, we standardize \(d_i\) and reduce it to 64 dimensions using PCA. We then fit a logistic probe whose decision value defines the soft preference
target:
\begin{equation}
t_i =
w^\top
\operatorname{PCA}
\left(
\operatorname{Std}(d_i)
\right) + b
\label{eq:preference-target}
\end{equation} A positive target $t_i$ favors Image memory, whereas a negative
target favors Text memory.

\subsection{Preference-Supervised Router Training}
\label{sec:router-training}

To enable modality selection, we train a pairwise router using the continuous preference targets derived in Section~\ref{sec:preference-target} as supervision. Formally, given interaction step $i$, let $s_i$ denote the final-token hidden state obtained from the actor prompt containing the current instruction and screenshot. For each modality $z \in \{T, I\}$, let $e_i^z$ denote the embedding of the corresponding retrieved memory candidate.

We project \(s_i\) and \(e_i^z\) to a shared dimension and apply
\(\ell_2\) normalization, yielding \(\hat{s}_i\) and \(\hat{e}_i^z\),
respectively. For each modality, we construct the candidate feature
\begin{equation}
v_i^z =
[\hat{s}_i;
\hat{e}_i^z;
\hat{s}_i \odot \hat{e}_i^z;
|\hat{s}_i-\hat{e}_i^z|],
\label{eq:router-feature}
\end{equation}
where \(\odot\) denotes element-wise multiplication.

A shared MLP scorer \(f_\theta\), parameterized by \(\theta\), maps each
candidate feature to a scalar score:
\begin{equation}
q_i^z = f_\theta(v_i^z).
\end{equation}
The router then computes the relative modality-preference score as
\begin{equation}
r_i = q_i^I - q_i^T.
\label{eq:router-score}
\end{equation}
Because the same scorer is applied to both modalities, \(q_i^T\) and \(q_i^I\) are directly comparable, and their difference \(r_i\) represents the relative preference for Image over Text memory.

We train the router to regress the soft preference target \(t_i\) using
\begin{equation}
\mathcal{L}_{\mathrm{route}}
=
\frac{1}{|\mathcal{D}_{\mathrm{pref}}|}
\sum_{i\in\mathcal{D}_{\mathrm{pref}}}
(r_i-t_i)^2.
\label{eq:router-loss}
\end{equation}
At inference, \(r_i\) is compared
with a predefined threshold \(\tau\). Image memory is
selected when \(r_i \geq \tau\); otherwise, Text memory is selected. Only the
selected memory is then provided to the frozen actor for action generation.

\section{Experiments}
\label{sec:experiments}

\paragraph{Benchmarks.}
We evaluate SwitchMem on three GUI benchmarks: MobileMiniWob++~\citep{rawles2025androidworld}, AITW~\citep{rawles2023android}, and LearnGUI~\citep{liu-etal-2026-learnact}. Together, these benchmarks cover diverse mobile GUI interaction settings and allow us to examine whether adaptive memory modality selection generalizes across environments. Because MobileMiniWoB++ does not provide official training or validation splits, we transfer the router trained on
LearnGUI while keeping its weights fixed. Further details on the transfer and threshold-calibration protocol are provided in
Section~\ref{sec:switchmem-training-details}.

\paragraph{Models and Retrieval.}
All primary experiments use Qwen3.5-9B~\citep{qwen3.5} as the frozen actor and
Qwen3-VL-Embedding-2B~\citep{li2026qwen3vlembedding} for top-1 retrieval and memory embedding. Text and Image
memories are instantiated from the same retrieved demonstration, isolating the
effect of memory representation.

\paragraph{Baselines.}
We compare SwitchMem with general-purpose VLMs (GPT-5.4-mini and Qwen3.5-9B),
GUI specialized agents, such as Aguvis \citep{xu2024aguvis} and UI-TARS variants \citep{qin2025ui}, and memory-augmented
methods. The memory-augmented group uses the same frozen Qwen3.5-9B actor and
includes RAG-GUI~\citep{xu2025retrieval},
M3A~\citep{rawles2025androidworld},
AppAgent~\citep{zhang2025appagent},
EchoTrail~\citep{li2026echotrail},
AGMem~\citep{choi2026naive},
and fixed Text, Image, and Both baselines.
\begin{table}[t]
\centering
\small
\setlength{\tabcolsep}{5pt}
\begin{tabular}{lrrrr}
\toprule
Benchmark
& Both 
& Image 
& Switches
& Run
\\
\midrule
AITW          & 83.8 & 36.5 & 1.72 & 2.42  \\
LearnGUI      & 90.8 & 81.3 & 1.79 & 4.95  \\
MobileMiniWoB & 57.8 & 76.3 & 1.56 & 4.50  \\
\bottomrule
\end{tabular}
\caption{
Within-trajectory routing behavior of SwitchMem.
\textbf{Both} and \textbf{Image} denote the percentages of dual-modality trajectories and Image-routed steps, respectively.
\textbf{Switches} and \textbf{Run} measure the average modality transition count and contiguous run length per trajectory.
}
\label{tab:trajectory-switching}
\end{table}

\subsection{Results}

\paragraph{Main results.}
Table~\ref{tab:results} shows that augmenting the GUI agent with
retrieved experience can improve action prediction, but its effectiveness depends on how the experience is represented. Image memory tends to perform well on AITW and LearnGUI, whereas providing both modalities is more effective on MobileMiniWoB. Notably, advanced memory formulation strategies do not consistently outperform raw context injection, suggesting that altering context representation without step-wise selectivity yields limited benefits.

SwitchMem achieves the strongest performance across all three benchmarks.
Rather than fixing one representation or always providing both, it selects a
single memory modality at each interaction step. This allows the actor to
exploit the complementary strengths of textual and visual memory across steps
while avoiding the performance degradation that can arise from jointly
providing both modalities.

\paragraph{Modality preference changes within trajectories.}
Table~\ref{tab:trajectory-switching} offers a quantitative analysis of SwitchMem's routing behavior across steps within a trajectory. Across all benchmarks, SwitchMem frequently utilizes both modalities within a single trajectory exhibiting moderate transition frequencies. Crucially, selected modalities are retained for consecutive steps. This demonstrates that SwitchMem does not lock into a static task level modality, but exhibits state adaptive modality.

\section{Analysis}
\subsection{Modality Utility Patterns and Adaptive Routing Alignment}
\label{sec:modality-dynamics}
\begin{figure}[t]
    \centering
    \includegraphics[width=0.92\linewidth]{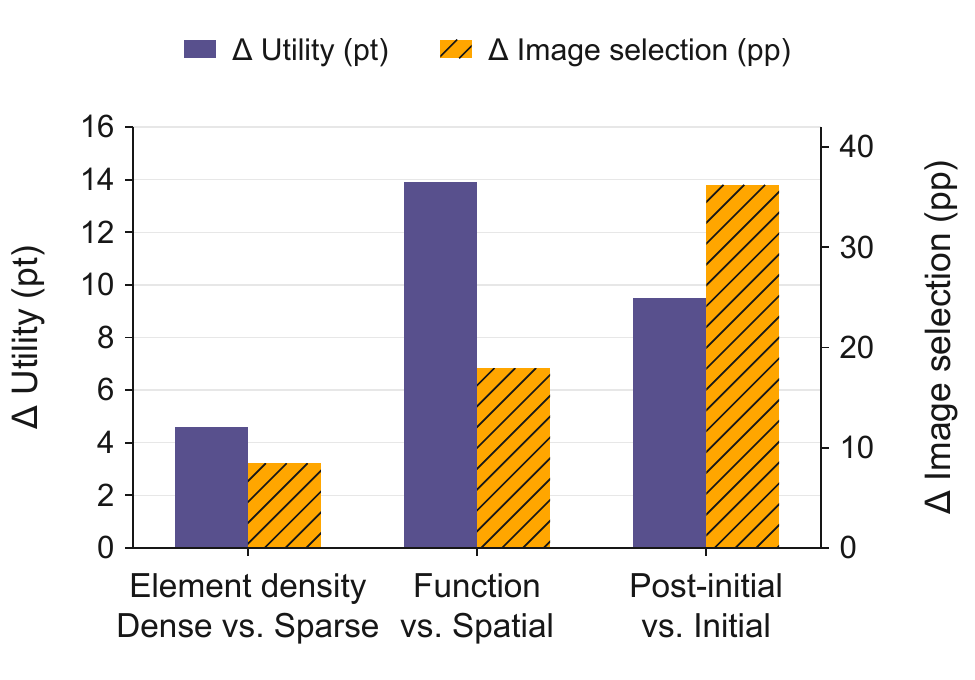}
    \caption{
    Changes in Image over Text utility (left axis) and SwitchMem's
    Image-selection rate (right axis).
    }
        \label{fig:utility-routing-alignment}
\end{figure}

To uncover the mechanisms governing dynamic modality selection, we analyze interaction context from two complementary perspectives. 
First, we identify when Image modality is preferred over Text discussed in Section ~\ref{sec:step-wise-preference}. 
Second, we evaluate whether SwitchMem's learned router aligns with these utility patterns to drive overall performance gains. Figure~\ref{fig:utility-routing-alignment} illustrates how SwitchMem's routing decisions align with these utility shifts across each dimension.
\subsubsection{When Is Each Memory Modality Useful?}
\label{sec:modality-usefulness}
\paragraph{Visually dense states favor Image memory.}
To quantify structural complexity, we measure \textbf{element density} using bounding box annotations provided in each benchmark. Specifically, we define element density as the number of unique UI elements divided by their fractional screen coverage, computed as the union area of the bounding boxes relative to the total screen area. Empirical analysis shows that element density remains positively associated with the Image-over-Text utility gap. Image memory yields an advantage when decisions rely on fine-grained layouts containing dense UI elements that textual abstractions fail to convey.
\paragraph{Image memory is more useful for function-oriented actions.}
Following standard GUI action taxonomies~\citep{shi2025mobilegui,Lu_2025_ICCV}, we group interaction steps into Spatial and Function actions. Spatial actions comprise coordinate-based operations such as tapping, swiping, scrolling, and dragging, whereas Function actions govern interface functionality, task status, or system navigation. When shifting from Spatial to Function actions, the utility advantage of Image memory grows  across benchmarks, showing that functional interface components rely on visual grounding. This suggests that visual memory is particularly valuable for
decisions requiring an understanding of interface state and
functionality, beyond simply locating interaction targets.
\paragraph{Image memory becomes more useful after the interface evolves.}
Compared to initial states, post-initial states exhibit a wider Image-over-Text utility gap. This finding indicates that as a trajectory progresses beyond initial states, evolving UI contexts generate complex visual nuances that text-based representations fail to preserve.

\begin{figure*}[t]
    \centering
    \begin{subfigure}[b]{0.32\textwidth}
        \centering
        \includegraphics[width=\linewidth]{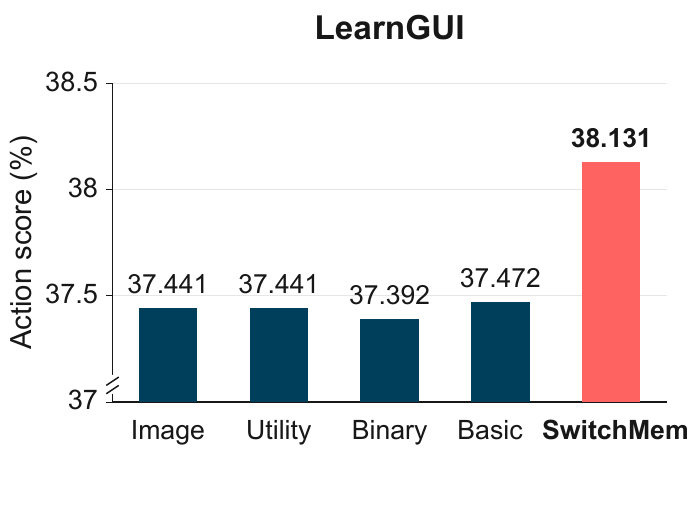}
        \caption{Routing baseline ablation}
        \label{fig:routing-baselines-learngui}
    \end{subfigure}
    \hfill
    \begin{subfigure}[b]{0.32\textwidth}
        \centering
        \includegraphics[width=\linewidth]{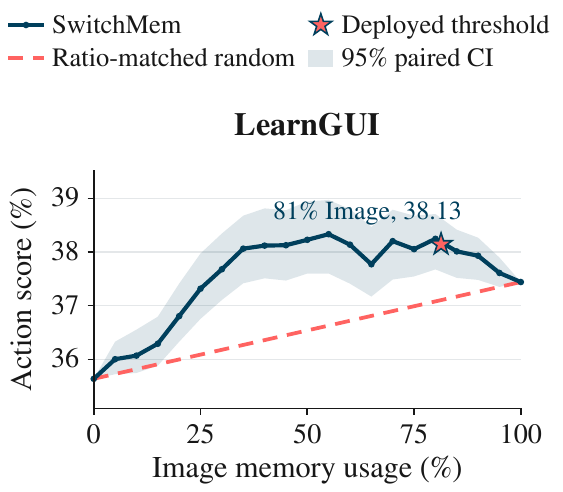}
        \caption{Modality budget frontier}
        \label{fig:routing-baseline}
    \end{subfigure}
    \hfill
    \begin{subfigure}[b]{0.32\textwidth}
        \centering
        \includegraphics[width=\linewidth]{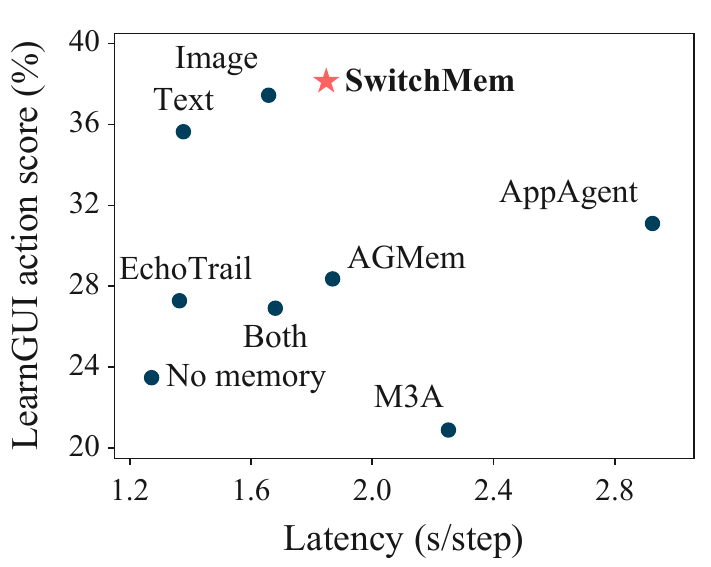}
        \caption{Latency--performance trade-off}
        \label{fig:latency}
    \end{subfigure}
    \caption{
        Empirical analysis on LearnGUI. 
        (a) Comparison with simpler routing baselines. Utility and Binary use direct preference supervision, while Basic routing excludes actor hidden states. 
        (b) SwitchMem versus ratio-matched random routing, where the dashed line denotes the analytic expected score and the star marks the deployed threshold. 
        (c) Efficiency Analysis of SwitchMem. RAG-GUI is omitted in figure due to its high latency of 11.118 s/step.
    }
    \label{fig:learngui-analysis}
\end{figure*}
\subsubsection{Understanding SwitchMem's Memory Modality Selection}
\label{sec:memory-modality-selection}
While Section~\ref{sec:modality-usefulness} identified when visual context is useful, we now examine whether SwitchMem's learned router dynamically aligns with these utility patterns. Furthermore, we evaluate how SwitchMem's adaptive routing translates into overall performance gains across benchmarks.

\paragraph{SwitchMem follows modality-utility patterns.}
SwitchMem's routing selections consistently align with the utility shifts identified across all three dimensions (Figure~\ref{fig:utility-routing-alignment}). The router selects Image memory more frequently on visually dense screens, at steps whose reference actions are categorized as Function rather than Spatial, and in post-initial states. This alignment demonstrates that SwitchMem's router dynamically responds to systematic changes in modality utility across interaction steps. This behavior helps explain the performance improvements in
Table~\ref{tab:results}. By favoring Image in conditions where visual
context is more useful while retaining Text in complementary states, SwitchMem
outperforms static memory strategies across all three benchmarks.

\subsection{Comparison with Alternative Routing Strategies}
\label{sec:preference-target-ablation}

To demonstrate the advantages of SwitchMem's routing design, we analyze its performance against alternative routing baselines.

\paragraph{Learned routing baselines.}
We compare SwitchMem with three learned alternatives.
\textit{Utility} and \textit{Binary} utilize the same router architecture as SwitchMem, constructed upon the actor's current-state hidden representation and retrieved-memory embeddings. However, both are trained directly
from sampled action utilities, without D-CoE-based target construction.
Utility regresses the utility gap using MSE, whereas Binary predicts
only its sign using binary cross-entropy.
\textit{Basic routing} also learns binary preferences, but uses embeddings
of the current instruction, screenshot, and retrieved memories rather
than the actor's hidden states. As shown in Figure~\ref{fig:routing-baselines-learngui}, conventional routing baselines offer minimal gains over a static Image Memory policy, whereas SwitchMem delivers consistent performance improvements. These results validate the necessity of actor-derived preference targets for guiding memory selection.

\paragraph{Ratio-matched random routing.}
To determine whether SwitchMem's gains arise from step-specific
memory selection rather than the overall Text and Image usage ratio,
we sweep its threshold and compare it with a random policy that
independently selects Image memory with probability $p$, matched
to SwitchMem's Image-selection rate.

Using the step level action scores under fixed Text and Image
memory, we compute the random policy's expected performance as
\begin{equation}
    S_{\mathrm{rand}}(p)
    =
    (1-p)S_{\mathrm{Text}} + pS_{\mathrm{Image}},
\end{equation}
where $S_{\mathrm{Text}}$ and $S_{\mathrm{Image}}$ denote the mean
action scores under the two fixed memory policies. Figure~\ref{fig:routing-baseline} shows that SwitchMem outperforms
ratio matched random routing at every evaluated intermediate
Image selection rate.
This indicates that its gains arise from assigning memory
modalities to appropriate interaction steps, rather than merely
using a particular Text and Image ratio.

\subsection{Efficiency Analysis}

Figure~\ref{fig:latency} compares the latency--performance trade off on
LearnGUI, with all methods measured on the same RTX PRO 6000 GPU.
SwitchMem achieves the highest action score while adding only modest latency. SwitchMem avoids processing both memory-conditioned prompts and incurs no additional autoregressive decision stage. In contrast, procedural baselines such as M3A, AppAgent, and RAG-GUI introduce additional annotation, document construction, or guideline-generation stages, resulting in higher latency despite lower performance. 


\begin{table}[t]
\centering
\normalsize
\renewcommand{\arraystretch}{1.12}
\resizebox{\columnwidth}{!}{
\begin{tabular}{lccc}
\toprule
Method & AITW  & LearnGUI  & MobileMiniWoB++  \\
\midrule
No memory & 29.980 & 25.705 & 23.913 \\
Text      & 32.640 & 38.230 & 32.609 \\
Image     & 30.440 & 40.997 & 33.696 \\
\rowcolor{switchblue}
SwitchMem & \textbf{33.680} & \textbf{41.055} & \textbf{36.957} \\
\bottomrule
\end{tabular}
}
\caption{
Generalization to the GUI specialized UI-TARS-7B-SFT actor.
SwitchMem achieves the best performance across all three benchmarks,
improving over the strongest fixed memory condition.
}
\label{tab:backbone-generalization}
\end{table}

\subsection{Generalization to a GUI specialized Actor}
To evaluate whether SwitchMem generalizes beyond its base actor, we apply its preference construction and routing protocol to the GUI specialized UI-TARS-7B-SFT model. As shown in Table~\ref{tab:backbone-generalization}, SwitchMem consistently outperforms static memory baselines across all three benchmarks. These results confirm that SwitchMem's routing formulation effectively transfers across distinct actor architectures and domain-specialized backbones.

\section{Conclusion}

We study memory modality as a step-dependent decision for VLM-based GUI agents.
Across benchmarks, neither Text nor Image memory consistently dominates,
and their utility varies across interaction steps. We show that memory modality preference is reflected at internal state through D-CoE analysis. Based on the representation-level signal, we introduce \textbf{SwitchMem} which uses modality conditioned representation shifts to
train a router that selects the suitable memory modality at each step.
SwitchMem improves over fixed modality and memory-augmentation baselines while
providing only one memory modality to the actor. Together, these results confirm that optimal memory selection is step-dependent, and that our SwitchMem approach can predict memory modality achieving performance gains over static alternatives.

\section*{Limitations}

Our evaluation focuses on mobile GUI benchmarks centered on Android
interfaces and controlled interaction settings. Generalization to iOS
interfaces and newer Android versions, which may introduce different
visual layouts and interaction conventions, remains unverified.

Our online evaluation uses emulated environments rather than physical
devices. Although these environments support controlled and reproducible
experiments, they do not fully capture device rendering differences,
system interruptions, network variability, or hardware constraints.
Evaluation on physical devices is an important direction for future work.

SwitchMem requires access to the actor's hidden states for both preference
target construction and routing at inference. It therefore cannot be
directly applied to models accessed through APIs that do not expose these
representations. Extending modality routing to such models remains an
open direction.

\bibliography{custom}

\newpage
\appendix
\section{Dataset and Baseline Details}
\label{sec:appendix-details}

\subsection{Dataset Details}
\label{sec:dataset-details}

We evaluate SwitchMem on three mobile GUI benchmarks that cover complementary
interaction settings: Android in the Wild (AITW), LearnGUI, and
MobileMiniWoB++.

\paragraph{Android in the Wild (AITW).}
AITW~\citep{rawles2023android} contains human-demonstrated Android interaction
trajectories spanning diverse applications and user goals.
Its heterogeneous interface states and multi-step trajectories make it suitable
for studying whether the preferred memory representation changes across
interaction steps.
We evaluate next-action prediction using the benchmark-specific action scorer.

\paragraph{LearnGUI.}
LearnGUI~\citep{liu-etal-2026-learnact} evaluates mobile GUI agents in demonstration augmented interaction settings, requiring agents to execute complex navigation tasks by conditioning on retrieved historical execution traces. Because agent performance hinges on effectively leveraging retrieved multimodal experience for dynamic step-wise decisions, it serves as an ideal testbed for evaluating adaptive memory routing. Memory is retrieved only from demonstration trajectories distinct from the query trajectory. Following the benchmark's standardized protocol, we report action level execution accuracy across its diverse mobile interaction tasks.

\paragraph{MobileMiniWoB++.}
MobileMiniWoB++~\citep{rawles2025androidworld} provides programmatically verifiable mobile GUI tasks in an online, closed-loop execution environment. While AITW and LearnGUI evaluate agents against heterogeneous real-world traces, MobileMiniWoB++ offers deterministic state transitions and immediate environment feedback for step-wise actions. This setup serves as a controlled testbed, allowing us to isolate and evaluate SwitchMem's dynamic memory routing without confounding noise from real-world OS rendering anomalies or ambiguous task rewards. Following standard protocol, we report task completion success rate.

Together, the three benchmarks cover human-demonstrated trajectories,
demonstration-based knowledge transfer, and controlled executable interaction.
For all benchmarks, Text and Image memory conditions use the same actor,
current observation, retrieved demonstration, and decoding configuration; only
the representation of the retrieved memory differs.
\subsection{Baseline Details}
\label{sec:baseline-details}

General-purpose and GUI specialized baselines are evaluated from their released
models or APIs. For a fair comparison, we standardize the backbone actor across all memory-augmented methods. Specifically, all methods use the same frozen Qwen3.5-9B actor, retrieval bank, and decoding protocol, differing only in how the retrieved experience is represented and provided to the actor.

\paragraph{Fixed memory conditions.}
\textsc{Text} serializes the retrieved instruction, state description, and
demonstrated action.  \textsc{Image} replaces the state description with its
screenshot while retaining the same instruction and action information.
\textsc{Both} supplies both representations of the same retrieved step.

\paragraph{RAG-GUI.}
\textsc{RAG-GUI}~\citep{xu2025retrieval} augments the agent with external procedural knowledge retrieved from a corpus of GUI tutorials. Rather than appending raw retrieved documents, it filters and converts relevant content into concise, task- and progress-conditioned textual guidelines for the actor during action prediction.

\paragraph{M3A.}
\textsc{M3A}~\citep{rawles2025androidworld} enriches the current observation with structured interface information, including visible UI elements and recent interaction history. This structured context enables the actor to jointly reason about screen states and prior execution steps during action selection.

\paragraph{AppAgent.}
\textsc{AppAgent}~\citep{zhang2025appagent} constructs an offline operation document by exploring target applications to detail interface elements, executable actions, and state transitions. At inference time, it retrieves documentation associated with the current screen’s UI elements and uses it to guide action generation.

\paragraph{EchoTrail.}
\textsc{EchoTrail}~\citep{li2026echotrail} builds reusable memory from collected GUI trajectories using self-exploration and critic-based filtering. Successful experiences are reformulated into compact textual traces highlighting interface states and transitions, enabling the retrieval and reuse of relevant interaction patterns for new tasks.

\paragraph{AGMem.}
\textsc{AGMem}~\citep{choi2026naive} preserves action-relevant visual evidence from demonstrations by comparing consecutive frames to crop changed interaction regions. It provides these cropped regions as visual memory, falling back to full demonstration frames when distinct regions cannot be reliably identified.

\vspace{1ex}
To ensure fair comparison, all baselines are implemented as training-free, component-level memory methods under the same retrieval bank, frozen actor, and decoding protocol as \textsc{SwitchMem}. Consequently, performance differences directly reflect how each method formulates and delivers retrieved experience to the agent.
\section{SwitchMem Details}
\subsection{SwitchMem Training and Evaluation Details}
\label{sec:switchmem-training-details}

For AITW and LearnGUI, we retrieve one memory step using the text-based
retriever and represent the same retrieved ID as either Text or Image. At each training state, we sample seven actor actions per
modality using temperature 0.7 and seeds 42--48, and score them with the
benchmark specific action metric. Exact utility ties are excluded from
preference target construction. The routing threshold
is selected by maximizing utility on the resulting training predictions and is frozen before test evaluation. The router is parameterized by a two-layer MLP ($\text{1024} \rightarrow \text{256} \rightarrow \text{1}$) with GELU activation and a $0.1$ dropout rate, trained using AdamW with a learning rate of $0.001$. 

\paragraph{AITW.}
We divide 20,000 steps from the official training split at the trajectory level
into 10,019 preference-query steps and 9,981 memory steps, corresponding to
3,100 and 3,087 trajectories, respectively. The two subsets have no trajectory
overlap. Each training query retrieves a demonstration from the memory subset
within the same AITW task category. For test evaluation, we use 5,000 steps from
759 held-out trajectories and allow retrieval from the complete 20,000-step
training bank.

\paragraph{LearnGUI.}
We use the 101 successful Online demonstration trajectories, comprising 1,500
interaction steps, as the source of both preference queries and candidate
memories. Success is used  to ensure that the memory bank contains
validated experiences.  We then construct preference
supervision by comparing the actor's action quality under the Text and Image
representations of the same retrieved step. At evaluation, we follow LearnGUI's
offline one-shot setting and use the support demonstrations available as memory in that
setting. The router and threshold remain fixed after Online training, and no
Offline-Test action, utility, or outcome is used for fitting or calibration.

\paragraph{MobileMiniWoB++.}
MobileMiniWoB++ does not provide step level action utilities or an official
training/validation split. We therefore transfer the LearnGUI router without
updating its weights. The memory bank is constructed once using suite seed
1001 and contains 129 interaction steps from 50 successful trajectories. It is
fixed before calibration and evaluation. For threshold calibration, we use
only the initial states of the 92 task templates instantiated with suite seed
43. Without observing actions, rewards, or task success, we choose the cutoff
whose Image-selection rate best matches the training-side routing rate on
LearnGUI. The cutoff is then frozen and evaluated through full online rollouts
on the same 92 templates instantiated with suite seed 42, yielding
63.043\% task success (58/92).

Suite seeds 1001, 43, and 42 share task templates but instantiate different task
values, on-screen text, and target coordinates. Consequently, the memory bank,
calibration set, and evaluation set contain no identical task instances. No
observation from seed 42 is used to construct the memory bank, train the router,
or select its threshold.

Text memory is constructed without an additional captioning model. For AITW,
we serialize the dataset-provided UI text annotations together with the current
Android activity. For LearnGUI, we use the provided element annotations,
including page captions, element descriptions, and functionality labels. For
MobileMiniWoB++, we deterministically serialize the Android accessibility
hierarchy obtained through UIAutomator. Each textual GUI observation is paired
with its task instruction and demonstrated action to form a Text memory entry.

\subsection{Modality-preference Probing Details}
\label{sec:modality-probing-details}

For AITW and LearnGUI and each layer
$\ell\in\{0,8,16,23,31\}$, we compute the signed
D-CoE shift $h^{\mathrm I}_{i,\ell}-h^{\mathrm T}_{i,\ell}$ from the final
 prompt token immediately before action generation.  Features are standardized, reduced to 64 PCA dimensions, and
classified with a logistic probe.  The label is retained only
when at least five of the seven sampled generations agree on the preferred
modality.

\begin{figure}[t]
    \centering \includegraphics[width=\linewidth]{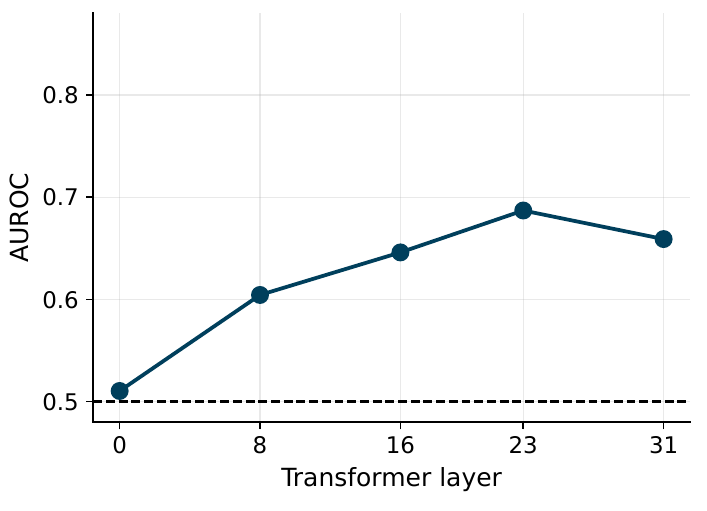}
    \caption{layer-wise modality-preference probing on Learn-GUI.  AUROC is computed by trajectory-grouped prediction
    on the 5-of-7 stable-preference subset.  The dashed line denotes chance.}
    \label{fig:appendix-aitw-layer-probe}
\end{figure}

The modality-preference signal becomes most pronounced in the later
transformer layers. It reaches its highest AUROC at layer 31 on AITW and at
layer 23 on LearnGUI. Rather than selecting the best layer separately for each
benchmark, SwitchMem consistently uses layer 31, the final transformer block.
Its hidden state represents the actor's fully integrated multimodal
representation immediately before action generation. This choice provides a
single benchmark-independent extraction rule and avoids tuning the
representation layer to each evaluation dataset, despite the higher
layer-23 probe score on LearnGUI.

We do not report an analogous 5-of-7 probe for MobileMiniWoB++ because the
benchmark does not provide the intermediate step level utilities required to
construct Text and Image preference targets. We therefore evaluate
MobileMiniWoB++ under a transfer setting using the router trained on LearnGUI,
which precludes constructing the corresponding utility-based probing subset.

\subsection{Additional Routing Strategy Comparisons}
\label{app:routing-baselines}

\begin{figure}[t]
    \centering
    \includegraphics[width=\linewidth]
    {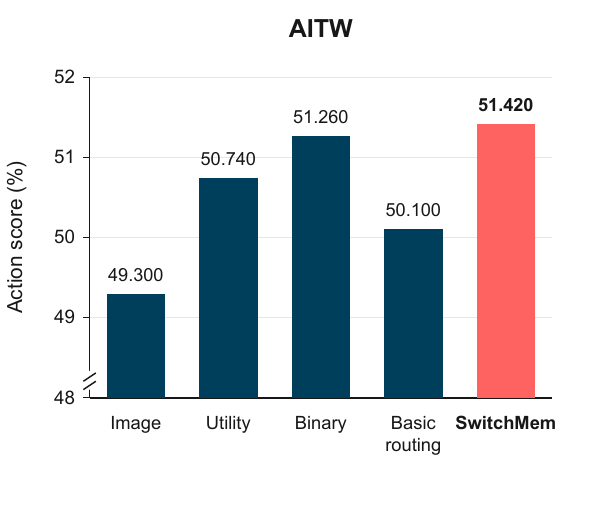}
    \caption{
    Comparison with simpler routing baselines on AITW.
    Binary is closely competitive with SwitchMem.
    The y-axis is truncated.
    }
    \label{fig:routing-baselines-aitw}
\end{figure}

We repeat the routing comparisons on AITW using the same
experimental protocol.
Figure~\ref{fig:routing-baselines-aitw} shows that Utility,
Binary, and Basic routing improve over the fixed Image policy,
but remain below SwitchMem.
Binary is particularly competitive on AITW, whereas the
advantage of SwitchMem is more pronounced on LearnGUI.
Across both benchmarks, SwitchMem achieves the highest point
estimates, supporting D-CoE-based target construction as a
more consistently effective alternative to direct supervision.

\begin{figure}[t]
    \centering
    \includegraphics[width=\linewidth]
    {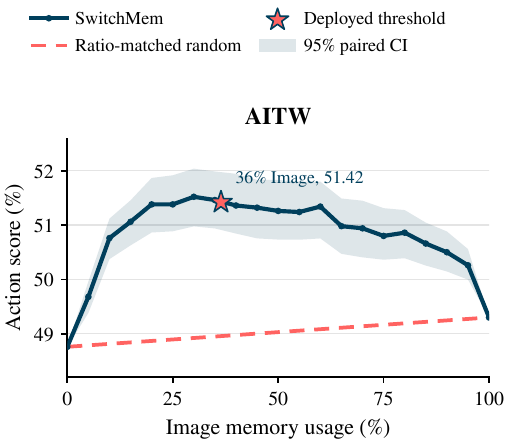}
    \caption{
    SwitchMem versus random routing on AITW.
    Random routing controls for aggregate Image usage but
    does not condition its selections on the interaction state.
    }
    \label{fig:routing-baseline-aitw}
\end{figure}

Figure~\ref{fig:routing-baseline-aitw} further shows that SwitchMem
outperforms ratio-matched random routing at every evaluated
non-trivial modality budget.
The same pattern on AITW and LearnGUI supports step-specific
modality assignment, rather than the global Text--Image ratio,
as a source of the adaptive-routing benefit.

\subsection{Efficiency Details}
\label{sec:appendix-efficiency-details}

Table~\ref{tab:appendix-latency} reports mean wall-clock latency on the same RTX
PRO 6000 GPU.  Requests are executed sequentially with greedy
decoding, thinking disabled, and the same 128-token output cap.  The
\textsc{SwitchMem} number includes its current-state hidden-state probe, the
small pairwise router, and one selected-memory action call.
Although \textsc{SwitchMem} introduces an additional routing stage, its runtime
overhead remains modest. Compared with the Image baseline, \textsc{SwitchMem}
adds only 0.128\,s, 0.190\,s, and 0.304\,s per step on AITW, LearnGUI, and
MobileMiniWoB++, respectively, while achieving consistently better action
performance.
In contrast, more elaborate memory-augmented baselines such as M3A, AppAgent,
and RAG-GUI introduce additional inference-time processing and incur
substantially higher latency, while still underperforming \textsc{SwitchMem}.
For example, RAG-GUI requires 8.815\,s and 11.118\,s per step on AITW and
LearnGUI, compared with 1.407\,s and 1.848\,s for \textsc{SwitchMem}.
Overall, these results show that adaptive memory routing provides consistent
performance gains with only a small additional inference cost.

\begin{table}[t]

\centering
\small
\setlength{\tabcolsep}{8pt}
\begin{tabular}{lrrr}
\toprule
Method & AITW & LearnGUI & MobileMini \\
\midrule
No memory & 1.297 & 1.272 & 0.906 \\
Text & 1.154 & 1.377 & 0.889 \\
Image & 1.279 & 1.658 & 1.558 \\
Both & 1.309 & 1.680 & 1.441 \\
EchoTrail & 1.205 & 1.364 & 1.189 \\
AGMem & 1.372 & 1.869 & 1.388 \\
M3A & 2.156 & 2.251 & 3.952 \\
AppAgent & 3.298 & 2.924 & 2.682 \\
RAG-GUI & 8.815 & 11.118 & 2.862 \\
\textbf{SwitchMem} & \textbf{1.407} & \textbf{1.848} & \textbf{1.862} \\
\bottomrule
\end{tabular}
\caption{Mean inference latency in seconds per interaction step.
 Method-specific inference stages
are included, while retrieval and input preprocessing are excluded.}
\label{tab:appendix-latency}
\end{table}

\subsection{Text and Image Memory Examples}

\begin{figure*}[t]
  \centering
  \begingroup
  \setlength{\parskip}{0pt}
  \setlength{\fboxsep}{1pt}
  \tcbset{
    equal height group=booking-memory,
    valign=top,
    boxsep=1mm,
    left=1.5mm,
    right=1.5mm,
    top=1mm,
    bottom=1mm,
    before skip=0pt,
    after skip=0pt
  }

  \begin{minipage}[t]{0.485\textwidth}
    \vspace{0pt}
    \begin{switchmemprompt}{(a) Text-memory context}
      \footnotesize
      \raggedright

      \memorylabel{Demonstration Memory}

      \textbf{Past task:}
      Open Booking.com. Look up a one-way flight from
      Melbourne to Auckland. Depart on June 22. 3 adults in business class.
      Stopovers are allowed. Sort by cheapest price.\par
      \vspace{0.3ex}

      \textbf{Past GUI observation:}
      This is a Booking.com app page where users
      can search for stays, flights, car rentals, and taxis. The page also shows
      loyalty program information and previous search results. \texttt{|}
      Messages \texttt{|} Notifications \texttt{|} Stays \texttt{|} Flights
      \texttt{|} Car rental \texttt{|} Taxi \texttt{|} Toronto \texttt{|}
      Staying from Sun, Sep 01 until Mon, Sep 30 \texttt{|} 1 room, 2 adults,
      0 children \texttt{|} Search \texttt{|} Jaeger, you're at Genius Level 1
      in our loyalty program \texttt{|} 10\% discounts on stays \texttt{|}
      Enjoy discounts at participating properties worldwide \texttt{|}
      1--30 Sep, 2 adults \texttt{|} Saved \texttt{|} Bookings \texttt{|}
      Profile \texttt{|} Accommodation search box \texttt{|} Travel more,
      spend less \texttt{|} Continue your search \texttt{|} Miami \texttt{|}
      10--20 Aug, 2 adults.\par
      \vspace{0.3ex}

      \textbf{Past action:}
      \texttt{\{"type":"tap",}\allowbreak
      \texttt{"position":[0.438194,0.121795]\}}\par
      \vspace{0.3ex}

      \textbf{Past action description:}
      On Booking.com Home Screen, click
      ``Flights'' button, to search for flights.
    \end{switchmemprompt}
  \end{minipage}
  \hfill%
  \begin{minipage}[t]{0.485\textwidth}
    \vspace{0pt}
    \begin{switchmemprompt}{(b) Image-memory context}
      \footnotesize
      \raggedright

      \memorylabel{Demonstration Memory}

      \textbf{Past task:}
      Open Booking.com. Look up a one-way flight from
      Melbourne to Auckland. Depart on June 22. 3 adults in business class.
      Stopovers are allowed. Sort by cheapest price.\par
      \vspace{0.3ex}

      \textbf{Past GUI observation:}\par
      \vspace{0.3ex}

      {\centering
        \fbox{%
          \includegraphics[
            height=4.0cm,
            width=0.48\linewidth,
            keepaspectratio
          ]{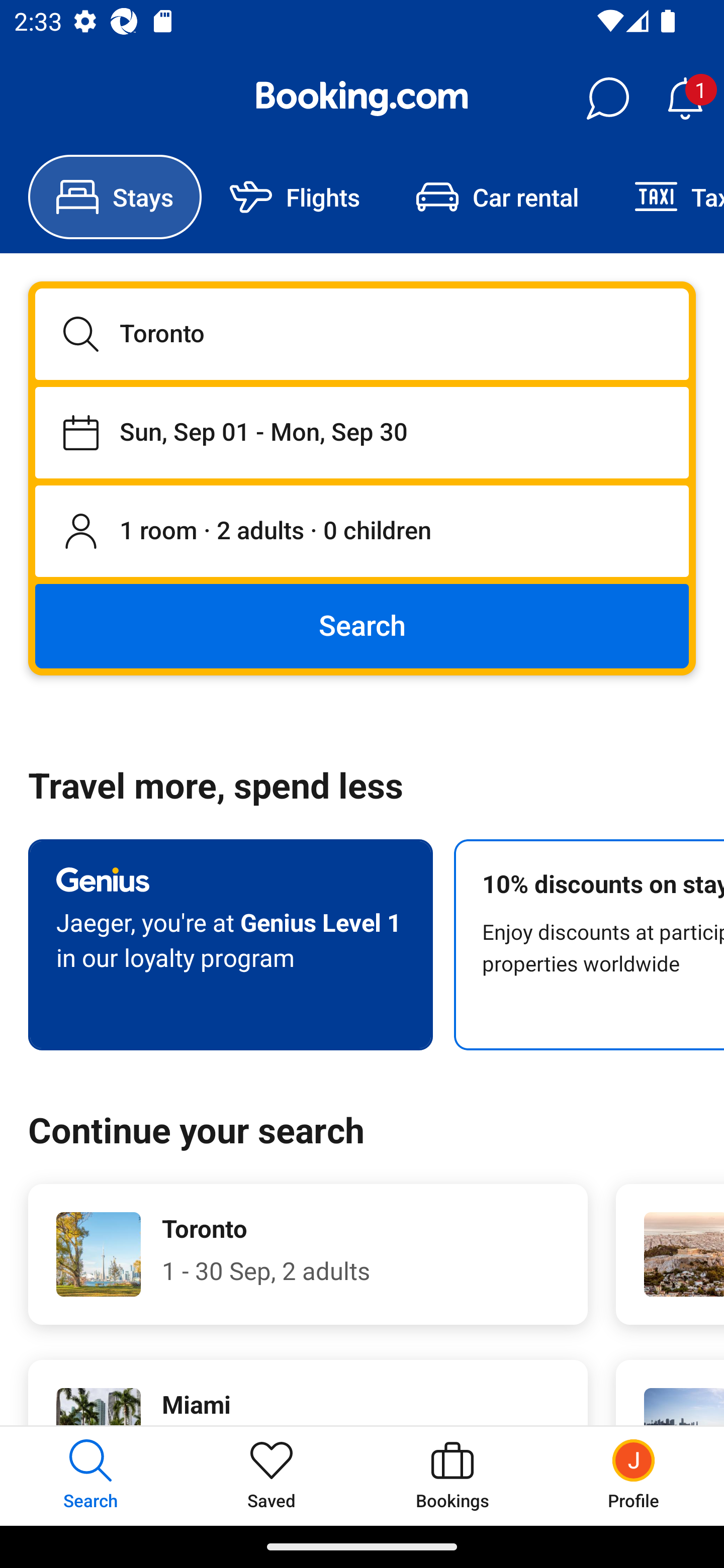}%
        }\par
      }
      \vspace{0.3ex}

      \textbf{Past action:}
      \texttt{\{"type":"tap",}\allowbreak
      \texttt{"position":[0.438194,0.121795]\}}\par
      \vspace{0.3ex}

      \textbf{Past action description:}
      On Booking.com Home Screen, click
      ``Flights'' button, to search for flights.
    \end{switchmemprompt}
  \end{minipage}

  \endgroup


  \caption{Paired actor prompts used by SwitchMem. Both variants contain the
  same retrieved step from a successfully completed trajectory and expose its
  demonstrated action. Text memory renders the past GUI state as a textual observation,
  whereas Image memory replaces only that observation with the corresponding full
  screenshot. The image is cropped only for visualization; the actor receives
  the full screenshot. The current task, current screenshot, action schema,
  actor, and decoding configuration are held fixed. For readability, the common system prompt and
  current screenshot are collapsed in the figure.}
  \label{fig:switchmem-prompt-example}
\end{figure*}

Figure~\ref{fig:switchmem-prompt-example} illustrates the text and
image memory representations used by SwitchMem. Both variants are constructed
from the same step retrieved from a successfully completed trajectory. They
contain the same past task, successful action, and natural-language action
description. The only difference is the representation of the past
observation: text memory provides a textual description of the GUI state,
whereas image memory provides the corresponding screenshot. This paired
construction isolates the effect of the observation modality while preserving
the demonstrated task and action.
\subsection{Statistical significance.}
We assess whether the improvement over the strongest fixed memory baseline is
statistically reliable using paired tests. For AITW and LearnGUI, we resample
complete task trajectories rather than individual steps, thereby preserving
within-trajectory dependence, and construct 95\% confidence intervals using
20,000 paired cluster-bootstrap replicates. We additionally perform a
two-sided cluster-level sign-flip randomization test. For MobileMiniWoB++, where
each task produces a binary success outcome, we use a paired task bootstrap and
an exact McNemar test.

\begin{table*}[t]
\centering
\small
\setlength{\tabcolsep}{4pt}
\begin{tabular}{lrrrr}
\toprule
Dataset & Best fixed & SwitchMem & $\Delta$ & 95\% CI \\
\midrule
AITW & 49.300 & 51.420 & +2.120 & [1.270, 2.945] \\
LearnGUI & 37.441 & 38.131 & +0.690 & [0.149, 1.263] \\
MobileMiniWoB++ & 60.870 & 63.043 & +2.174 & [-3.261, 7.609] \\
\bottomrule
\end{tabular}
\caption{Paired performance differences between SwitchMem and the strongest
fixed memory baseline. Confidence intervals are reported in percentage points.}
\label{tab:significance}
\end{table*}

The improvements are statistically significant on AITW ($p<.001$) and
LearnGUI ($p=.016$), and remain significant after Holm correction across the
three benchmarks. On MobileMiniWoB++, SwitchMem exhibits a positive gain, but
the difference is not statistically significant ($p=.688$), reflecting the
limited sample size of 92 evaluation tasks. Moreover, the router
is transferred directly from LearnGUI without any MobileMiniWoB++ preference
supervision or target-domain training. Therefore, this result should be
interpreted as evidence of cross-benchmark transfer rather than
in-domain-optimized performance.
\clearpage
\makeatletter
\global\@colroom\textheight
\global\vsize\textheight
\makeatother
\raggedbottom
\subsection{Case Studies of When Each Memory Modality Is Useful}
\label{sec:case-studies}

We illustrate state-dependent memory selection using two states from
different AITW tasks. The first is a Sparse initial state requiring a
Spatial action, whereas the second is a Dense post-initial state
requiring a Function action. These examples contrast the three
contextual factors examined in our modality utility analysis.

Within each example, Text and Image memory represent the same retrieved
demonstration, and the current instruction and screenshot are identical
across memory conditions. Only the representation of the past GUI
observation differs. 

\begin{figure*}[t]
  \centering
  \begingroup
  \setlength{\parskip}{0pt}
  \setlength{\fboxsep}{1pt}

  \tcbset{
    equal height group=case-bestbuy-memories,
    valign=top,
    boxsep=1mm,
    left=1.5mm,
    right=1.5mm,
    top=1mm,
    bottom=1mm,
    before skip=0pt,
    after skip=0pt,
    fonttitle=\footnotesize\bfseries
  }

  {\small
    \textbf{Current task:}
    Go to the official website of bestbuy.\par
  }
  \vspace{0.8ex}

  \begin{minipage}[t]{0.24\textwidth}
    \vspace{0pt}
    \centering
    \footnotesize

    \textbf{(a) Current state: Step 0}\par
    \vspace{0.6ex}

    \includegraphics[
      height=5.2cm,
      width=0.95\linewidth,
      keepaspectratio
    ]{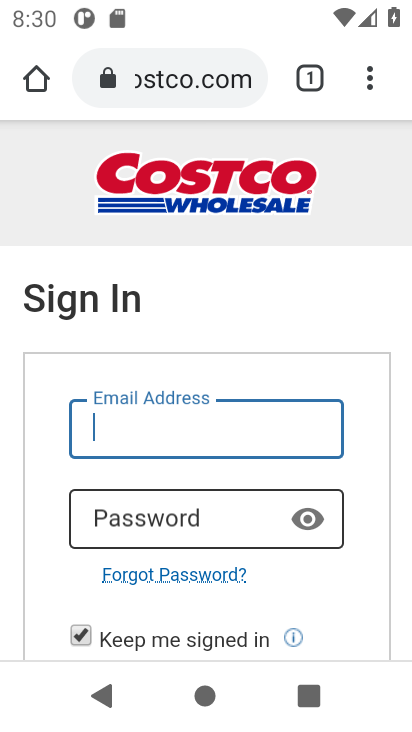}

    \par\vspace{0.6ex}
    Sparse / Spatial / Initial\par
    Element density: 177.25\par
    SwitchMem: \textbf{Text}\par
    Score (T/I): \textbf{1/0}
  \end{minipage}
  \hfill%
  \begin{minipage}[t]{0.375\textwidth}
    \vspace{0pt}
    \begin{switchmemprompt}{(b) Text memory}
      \footnotesize
      \raggedright

      \memorylabel{Demonstration Memory}

      \textbf{Past task:}
      open google chrome and search for bestbuy.com .\par
      \vspace{0.5ex}

      \textbf{Past GUI observation:}\par
      Android activity:
      \texttt{com.android.chrome/}\allowbreak
      \texttt{com.google.android.}\allowbreak
      \texttt{apps.chrome.Main}\par
      Visible UI text:
      \texttt{signin.costco.com/}\allowbreak
      \texttt{e0714dd4-}\allowbreak
      \texttt{784d-46d6-a278}
      \texttt{|} COSTCO
      \texttt{|} EWHOLESALE
      \texttt{|} Sign
      \texttt{|} In
      \texttt{|} -Email Address
      \texttt{|} Password
      \texttt{|} Eorgot.Password?
      \texttt{|} Keep me signed in
      \texttt{|} Check this
      \texttt{|} box only when
      \texttt{|} private device.
      \texttt{|} on
      \texttt{|} a
      \texttt{|} New to Costco?
      \texttt{|} Create Account
      \texttt{|} Terms and Conditions
      \texttt{|} Privacy Rights.\par
      \vspace{0.5ex}

      \textbf{Past action (normalized):}\par
      \texttt{\{"type":"tap",}\allowbreak
      \texttt{"position":}\allowbreak
      \texttt{[0.5313,0.0408]\}}
    \end{switchmemprompt}
  \end{minipage}
  \hfill%
  \begin{minipage}[t]{0.355\textwidth}
    \vspace{0pt}
    \begin{switchmemprompt}{(c) Image memory}
      \footnotesize
      \raggedright

      \memorylabel{Demonstration Memory}

      \textbf{Past task:}
      open google chrome and search for bestbuy.com .\par
      \vspace{0.5ex}

      \textbf{Past GUI observation:}\par
      \vspace{0.5ex}

      {\centering
        \fbox{%
          \includegraphics[
            height=4.0cm,
            width=0.85\linewidth,
            keepaspectratio
          ]{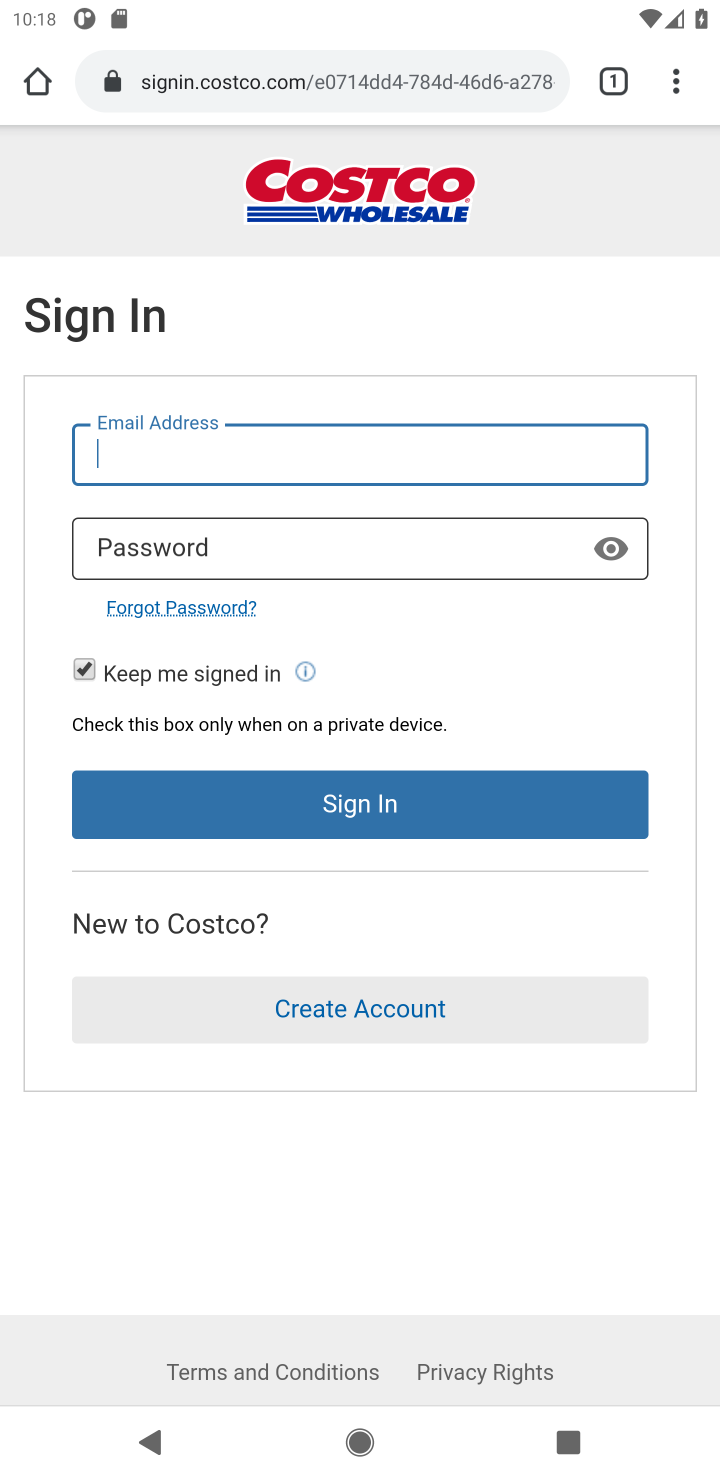}%
        }\par
      }
      \vspace{0.5ex}

      \textbf{Past action (normalized):}\par
      \texttt{\{"type":"tap",}\allowbreak
      \texttt{"position":}\allowbreak
      \texttt{[0.5313,0.0408]\}}
    \end{switchmemprompt}
  \end{minipage}

  \endgroup
  \caption{
    Text-favoring AITW case.
    (a) A Sparse initial state requiring a Spatial action.
    (b,c) Text and Image representations of the same retrieved
    demonstration, which also shows a Costco sign-in page.
    Text memory yields a correct tap on the current browser address
    bar, whereas Image memory yields a typing action before selecting
    the bar. SwitchMem selects Text.
    T/I denotes Text/Image action-matching scores;
    memory coordinates are normalized and rounded.
  }
  \label{fig:case-text}
\end{figure*}

\paragraph{Sparse initial state: Text memory is more useful.}
At step 0 of ``Go to the official website of bestbuy,'' the current
screen shows a Costco sign-in page
(Figure~\ref{fig:case-text}).
Its annotated element density is 177.25, below AITW's lower-quartile
cutoff of 310.00, placing it in the Sparse group.
The gold action is a tap on the browser address bar, classified
as Spatial. This state is therefore Sparse, Spatial, and Initial.

The retrieved demonstration has the instruction
``open google chrome and search for bestbuy.com'' and also depicts
a Costco sign-in page
(Figure~\ref{fig:case-text}).
Its textual observation includes the browser URL and labels such as
``Sign In,'' ``Email Address,'' and ``Password,'' together with the
demonstrated tap action. The Image condition instead provides the
screenshot of that past page, while retaining the same task and action
information.
Under Text memory, the actor correctly taps the current address bar.
Under Image memory, it outputs a typing action for
\texttt{bestbuy.com} without first selecting the address bar.
The action scores are 1 and 0, respectively, and SwitchMem selects Text.

\begin{figure*}[t]
  \centering
  \begingroup
  \setlength{\parskip}{0pt}
  \setlength{\fboxsep}{1pt}

  \tcbset{
    equal height group=case-walmart-memories,
    valign=top,
    boxsep=1mm,
    left=1.5mm,
    right=1.5mm,
    top=1mm,
    bottom=1mm,
    before skip=0pt,
    after skip=0pt,
    fonttitle=\footnotesize\bfseries
  }

  {\small
    \textbf{Current task:}
    Search for dell alienware on walmart.com.\par
  }
  \vspace{0.8ex}

  \begin{minipage}[t]{0.24\textwidth}
    \vspace{0pt}
    \centering
    \footnotesize

    \textbf{(a) Current state: Step 2}\par
    \vspace{0.6ex}

    \includegraphics[
      height=5.2cm,
      width=0.95\linewidth,
      keepaspectratio
    ]{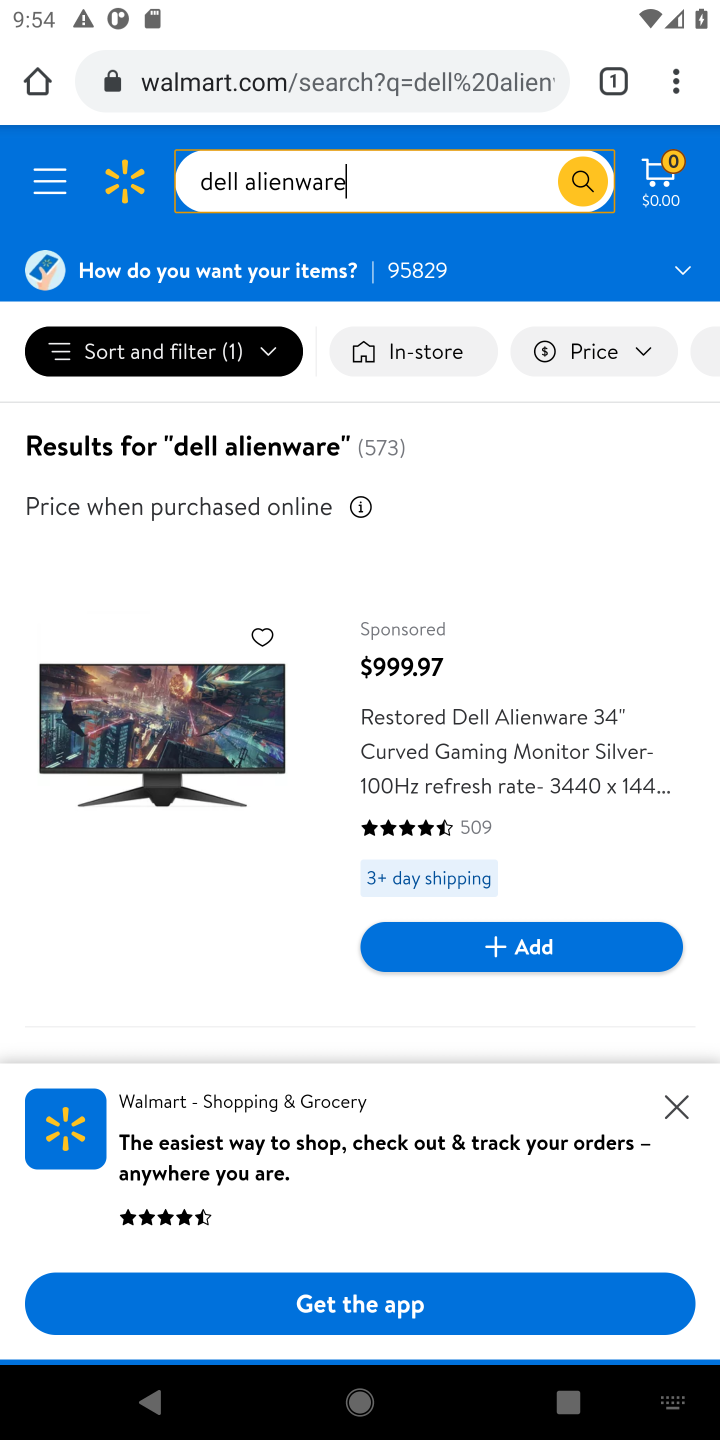}

    \par\vspace{0.6ex}
    Dense / Function / Post-initial\par
    Element density: 713.27\par
    SwitchMem: \textbf{Image}\par
    Score (T/I): \textbf{0/1}
  \end{minipage}
  \hfill%
  \begin{minipage}[t]{0.375\textwidth}
    \vspace{0pt}
    \begin{switchmemprompt}{(b) Text memory}
      \footnotesize
      \raggedright

      \memorylabel{Demonstration Memory}

      \textbf{Past task:}
      Search dell alienware on amazon.\par
      \vspace{0.5ex}

      \textbf{Past GUI observation:}\par
      Android activity:
      \texttt{com.android.chrome/}\allowbreak
      \texttt{com.google.android.}\allowbreak
      \texttt{apps.chrome.Main}\par
      Visible UI text:
      amazon.com
      \texttt{|} =
      \texttt{|} amazon
      \texttt{|} Sign
      \texttt{|} in
      \texttt{|} >
      \texttt{|} G
      \texttt{|} Define
      \texttt{|} Select
      \texttt{|} all
      \texttt{|} Copy
      \texttt{|} Share
      \texttt{|} Web search
      \texttt{|} X
      \texttt{|} dell alienware
      \texttt{|} bc
      \texttt{|} deul alienware aw3423dw
      \texttt{|} dell alienware aw3423dwf
      \texttt{|} dell alienware x14
      \texttt{|} dell alienware monitor
      \texttt{|} dell alienware x17
      \texttt{|} r2
      \texttt{|} essentials under \$50
      \texttt{|} TDAMEH
      \texttt{|} holiday gifts
      \texttt{|} Great
      \texttt{|} Be
      \texttt{|} Oculus
      \texttt{|} Women's Fashion
      \texttt{|} Pic
      \texttt{|} here
      \texttt{|} are
      \texttt{|} alienware.\par
      \vspace{0.5ex}

      \textbf{Past action (normalized):}\par
      \texttt{\{"type":"tap",}\allowbreak
      \texttt{"position":}\allowbreak
      \texttt{[0.1878,0.2123]\}}
    \end{switchmemprompt}
  \end{minipage}
  \hfill%
  \begin{minipage}[t]{0.355\textwidth}
    \vspace{0pt}
    \begin{switchmemprompt}{(c) Image memory}
      \footnotesize
      \raggedright

      \memorylabel{Demonstration Memory}

      \textbf{Past task:}
      Search dell alienware on amazon.\par
      \vspace{0.5ex}

      \textbf{Past GUI observation:}\par
      \vspace{0.5ex}

      {\centering
        \fbox{%
          \includegraphics[
            height=4.0cm,
            width=0.85\linewidth,
            keepaspectratio
          ]{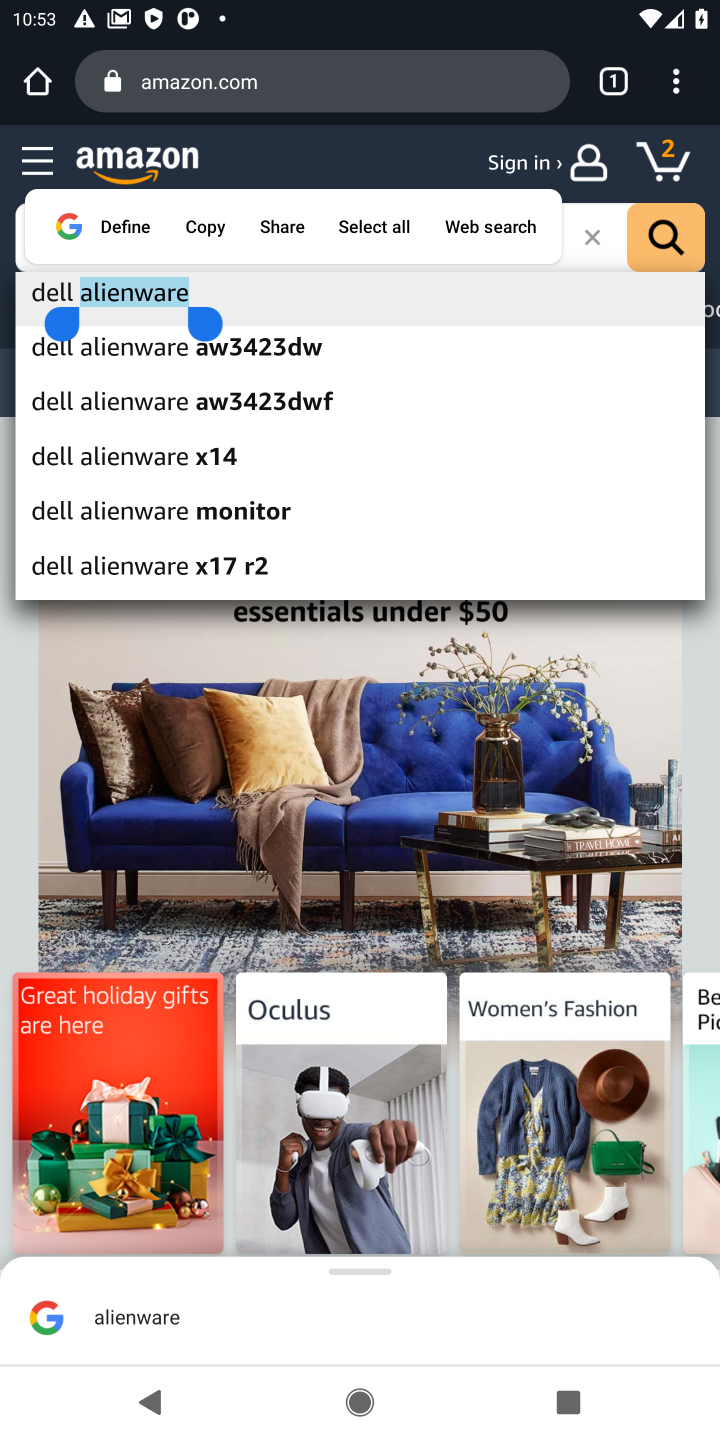}%
        }\par
      }
      \vspace{0.5ex}

      \textbf{Past action (normalized):}\par
      \texttt{\{"type":"tap",}\allowbreak
      \texttt{"position":}\allowbreak
      \texttt{[0.1878,0.2123]\}}
    \end{switchmemprompt}
  \end{minipage}

  \endgroup
  \caption{
    Image-favoring AITW case.
    (a) A Dense post-initial state requiring a Function action:
    the requested Walmart search results are already displayed.
    (b,c) Text and Image representations of the same retrieved
    Amazon demonstration at a query-entry state.
    Image memory yields correct task completion, whereas Text memory
    yields an unnecessary Enter action. SwitchMem selects Image.
    T/I denotes Text/Image action-matching scores;
    memory coordinates are normalized and rounded.
  }
  \label{fig:case-image}
\end{figure*}

\paragraph{Dense post-initial state: Image memory is more useful.}
At step 2 of ``Search for dell alienware on walmart.com,'' the current
screen displays the requested search results, including a query field,
filter controls, product information, and pricing
(Figure~\ref{fig:case-image}).
Its annotated element density is 713.27, above AITW's upper-quartile
cutoff of 567.73, placing it in the Dense group.
The gold action is \texttt{task\_complete}, categorized as
Function because it signals task completion rather than a spatial
interaction.
This state is therefore Dense, Function, and Post-initial.

The retrieved demonstration concerns ``Search dell alienware on amazon''
(Figure~\ref{fig:case-image}).
Its textual observation lists the query, search suggestions, and
text-selection controls, while its screenshot depicts the query-entry
state with the search text selected.
Under Text memory, the actor unnecessarily presses Enter.
Under Image memory, it correctly outputs \texttt{task\_complete}.
The action scores are 0 and 1, respectively, and SwitchMem selects Image.

Together, these examples illustrate contextual variation in memory
usefulness across interaction states. SwitchMem follows the
higher-scoring modality in both cases, although these illustrative
examples do not isolate the causal contribution of each contextual factor.

\end{document}

%% file: math_commands.tex
\usepackage{amsmath,amsfonts,bm}

\def\eqref#1{equation~\ref{#1}}

\def\1{\bm{1}}

\DeclareMathAlphabet{\mathsfit}{\encodingdefault}{\sfdefault}{m}{sl}
\SetMathAlphabet{\mathsfit}{bold}{\encodingdefault}{\sfdefault}{bx}{n}

